\documentclass[10pt]{article}

\usepackage[margin=1.3in]{geometry}  
\usepackage[parfill]{parskip}  
\usepackage{graphicx}
\usepackage{amsmath,amssymb,amsthm}
\usepackage{epstopdf}
\usepackage{xifthen}
\usepackage{enumitem}
\usepackage[colorlinks]{hyperref}
\usepackage{todonotes}
\usepackage{mathtools}
\usepackage[displaymath, mathlines]{lineno}
\usepackage{bm}

\usepackage{fancyhdr}
\usepackage{sectsty}
\sectionfont{\fontsize{10}{15}\selectfont}
\subsectionfont{\fontsize{10}{15}\selectfont}
\subsubsectionfont{\fontsize{10}{15}\selectfont}

\usepackage{hyperref}
\usepackage{tikz}
\usetikzlibrary{patterns,arrows,calc,decorations.pathmorphing}
\usetikzlibrary{matrix,chains,positioning,decorations.pathreplacing,arrows}
\usetikzlibrary{positioning,calc}
\usetikzlibrary{arrows,backgrounds}
\usetikzlibrary{angles,quotes}
\usetikzlibrary{cd}

\usepackage{setspace}

\usepackage{algorithm}
\usepackage{algpseudocode}

\DeclareMathAlphabet{\mathsfit}{\encodingdefault}{\sfdefault}{m}{sl}
\SetMathAlphabet{\mathsfit}{bold}{\encodingdefault}{\sfdefault}{bx}{n}

\def\gI{{\mathcal{I}}}

\usepackage{xargs}                      % Use more than one optional parameter in a new commands
\newcommand{\lc}{\left\{}
\newcommand{\rc}{\right\}}

\newcommand{\ds}{\mathrm{ds}}

\newcommand{\rep}{\stackrel{\text{rep.}}{\longrightarrow}}

\newcommand{\I}{\mathcal{I}}

\usepackage{amssymb,amsmath,amsthm, amsfonts}

\theoremstyle{plain}
\newtheorem{theorem}{Theorem}[section]

\theoremstyle{definition}

\theoremstyle{remark}
\newtheorem{remark}[theorem]{Remark}

\hypersetup{
    colorlinks=true,
    linkcolor=blue,
    filecolor=magenta,      
    urlcolor=cyan,
    pdftitle={Overleaf Example},
    pdfpagemode=FullScreen,
    citecolor = blue,
    }

\usepackage{subfigure}

\usepackage[style=numeric,backend=bibtex,maxbibnames=99]{biblatex}
\usepackage{caption}

\title{Characterizing Learning Dynamics under Relative Reparameterization of Singular Models}

\author{Pascal Mattia Esser\\Ludwig-Maximilians-Universit\"at M\"unchen, Germany,
\and
Frank Nielsen\\Sony Computer Science Laboratories Inc., Japan}

\begin{document}

\maketitle

\begin{abstract}
\noindent A common way to analyze learning of statistical models is to consider operations in the models parameter space, however this becomes challenging when there is no one-to-one mapping between the parameter space and the underlying statistical model space. Such ``singular models'' occur frequently and exhibit a characteristic decrease in convergence speed of learning trajectories due to attractor behaviors. In this work, we consider a relative reparameterization technique of the parameter space, which yields a general method for extracting regular sub-models from singular models. On the example of Gaussian Mixture Models and Neural Networks we theoretically and numerically analyze the convergence rate for Gradient Descent under both parameterizations. Analyzing second-order methods and explicit properties of the Fisher Information Matrix we distinguish between differences in convergence behavior arising from algorithmic and intrinsic information-geometric aspects.

\end{abstract}

\section{Introduction and Problem Setup}
A central focus of current machine learning research aims to understand the behavior of a model in the parameter space, for example, by studying optimizers and what local optima they converge to. the interdependence of the parameters and the underlying probability space that the model becomes especially interesting if \emph{the map from the parameter space to the probability distribution is not one-to-one}. We can formulate this class of machine learning of functions as follows.

\noindent\textbf{Singular Models.} Let a parametric regular statistical model $\mathcal{P}$ be a family of probability distributions $\mathcal{P}=\{p_\psi:\psi\in \Psi\}$ where $p_\psi$ is a function parameterized by $\psi$, an \emph{open} subset of $\mathbb{R}^K$. A statistical model is said \emph{identifiable} iff $p_\psi=p_{\psi'}\ \Leftrightarrow\ \psi=\psi'$. 
In information science such a parametric model is called \emph{singular}. In addition, a model $\{m_\psi(x)\ :\psi\in\Psi\}$ is said to be {\em identifiable} iff the mapping $\psi\mapsto m_\psi(x)$ is one-to-one. Furthermore a model is said {\em regular} if it is both identifiable and $\partial_1 m_\psi(x),\ldots, \partial_D m_\psi(x)$ are linearly independent. A regular model allows to define a manifold with tangent planes at $m_\psi$ using the basis $\{\partial_i m_\psi(x)\}$. In the vanilla case, a parametric statistical model is regular and identifiable, but even very simple examples violate this assumption. The study of singular models is a very relevant setting in practice as a lot of statistical models used in practice, such as  neural networks (NN), gaussian mixture models (GMM), reduced rank regressions, normal or binomial mixtures, hidden Markov models, stochastic context-free grammars and Bayesian networks are singular models. For all those  models, two main problems arise: firstly one can easily observe that all standard information criteria \cite{BIC,AIC,watanabe_2009} do no longer hold in this setting and secondly  singular points in the parameter space induce attractor behavior of the learning trajectories \cite{SingularDynamics, SingularDynamics2,SingularDynamica3, watanabe_2009, attractor}. In this work we focus on the second aspect.

\noindent\textbf{Learning Dynamics.} We consider the setting of learning the parameters of the model with a gradient descent (GD) by minimizing the objective function $\mathcal{L}(\Psi) = \frac{1}{n}\sum_{i=1}^n l(x;\psi)$ where $l(x;\Psi)$ is the log likelihood. Then the GD update at time step $t$ is defined as $\Psi^{(t+1)}=\Psi^{(t)}+\eta \nabla g^{-1}\mathcal{L}(\Psi^{(t)})$, where $\eta$ is the learning rate and $g$ a metric on the parameter space. The goal is to analyze how the parameters $\Psi^{(t)}$ evolve over time towards the true $\Psi^\star$. In a \emph{first order} setting $g=\mathbb{I}$ is the Euclidean metric and we recover the standard vanilla GD setting. In the \emph{second order} setup we assume $g=\gI(\Psi)$ to be the Fisher Information Matrix (FIM) \cite{NGD1}, which is a way of measuring the amount of information that an observable random variable $x$ carries about an unknown parameter $\psi$ of a distribution that models the samples $X$.

\noindent\textbf{Relative Reparameterization.}
By defining a reference parameter we construct a relative reparameterization of the model, that results in a new \emph{identifiable} model and our goal is to characterize the difference in the convergence of the learning dynamics.
We start with formalizing this notion: Consider a setting where $g(\cdot)$ is defined as a weighted sum of $f(\cdot)$ such that $g( X| \vartheta, \Phi) := \sum_{k=1}^K \theta^{(t)} f( X|\phi^{(t)})$, one can observe that if $\theta_i = \theta_j$ we cannot identify the parameters $\phi_i, \phi_j$.
Now let $\mathcal{P}=\{p_\theta:\theta\in \vartheta\}$, then a submodel as $\mathcal{S}\subset \mathcal{P}$ can be defined over a subset of the parameter space: $\mathcal{S}=\{p_\lambda:\Lambda^\prime\in \Lambda| \vartheta'\subseteq \vartheta\}$. Having defined the properties of the new parameter space we can now define a procedure to construct such a space from the original parameter space.

Consider an overlap singularity where we cannot identify $\phi_i, \phi_j$ iff $\theta_i = \theta_j$. Now starting from a parameterization $\vartheta = \lc \theta_1, \cdots \theta_{k}\rc$ and want to ensure an ordering of the form $\exists l\quad \theta_{\sigma(1)}^l< \theta_{\sigma(2)}^l<\cdots< \theta_{\sigma(K)}^l$ where $\sigma(\cdot)$ is a permutation that describes the parameter ordering after initialization. Wlog. let $\theta_{\sigma(1)}$ be the reference parameter. To enforce this define a set of \emph{relative parameters} over the difference $\{\Delta_i\}_{i=1}^{K-1}$ such that for any consecutive two parameters with $k\in \{2,\cdots,K\}$ $\xi:\exists l~ \theta^l_{\sigma(k)}\rep \theta^l_{\sigma(k-1)} +\Delta_{k-1}$ holds and therefore the set of parameters changes from $\vartheta := \lc \theta_1, \cdots \theta_{K+1}\rc$ to $\Lambda := \lc \theta_1, \Delta_1 \cdots\Delta_{K-1}\rc$. We refer to this as the \emph{reparameterized model}. If we additionally assume $\Delta_i\geq0\forall i \in[K]\ s.t.\ \exists l~ \theta_{\sigma(1)}^l\leq\cdots\leq \theta_{\sigma(K)}^l$ we refer to it as the \emph{ordered reparameterized model.}

\noindent\textbf{Contributions.} For GMMs we derive in Section~\ref{sec: GMM} convergence rate for vanilla, constraint and projected GD under standard parameterization and relative reparameterization, which allows us compare the behavior away, near and at the singularity. Extending our analysis we compare the convergence under the different parameterizations for second order methods. From there we show in Section~\ref{sec: NN} that the observations can be extended to commonly used models such as NNs as well. Finally in Section~\ref{sec: FIM} we analyze the geometry of the model through properties of the FIM and show that the differences in convergence arises from an algorithmic but not intrinsic viewpoint.

\section{Gaussian Mixture Models}\label{sec: GMM}

\begin{figure*}[t]
    \centering
    \includegraphics[width=\linewidth]{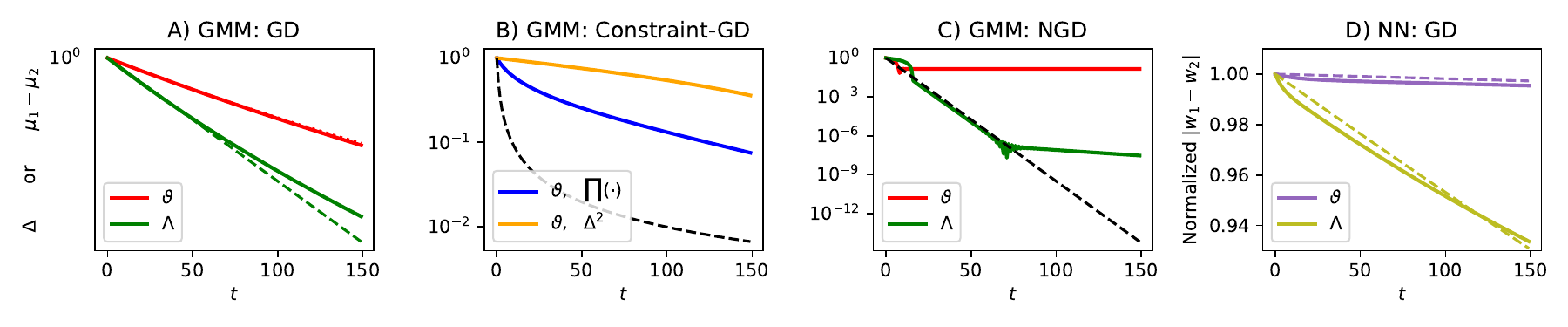}
    \caption{Illustration of the (normalized) empirical convergence rates (solid lines) and theoretical bounds (dashed lines) of Theorem~\ref{lem: GD standard}---\ref{lem: NN standard}.
    \textbf{A)-C)} For the GMM model we consider: $n=50000,T=150,\mu_1=-0.3,\mu_2=0.3$. $\rho$ is estimated from the first $20$ samples. We assume, the true model is singular with $\mu_1=\mu_2$. 
    \textbf{A)} Shows the convergence of the empirical and theoretical bound coincide well. 
    For \textbf{B)} we observe, that while the shape is roughly given by the theoretical bound the diverge. Importantly note that the Lemmas only give the order of the rate so the empirical difference is to be expected.
In \textbf{C)} we observe that for NGD, as expected both parameterizations match the overall bound however we observe that the $\Lambda$-parameterization stays numerically stable for longer before converging.
\textbf{D)} We simulate the NN model from section~\ref{sec: NN} and observe that for the $\Lambda$-parameterization the empirical convergence and theoretical bound match well. For the $\vartheta$-parameterization we estimate the constant $c$ from the late–time $\Theta$–dynamics by using the asymptotic update 
$\varepsilon_{t+1}-\varepsilon_t \approx -c\,\varepsilon_t^3$.   }
    \label{fig: convergence}
\end{figure*}

Consider a 2-GMM, defined as $p( X |\left\{ \pi_{k}, \mu_{k}, \sigma_{k}\right\}_{k=1}^2)=\pi_{1} \mathcal{N}\left( X | \mu_{1}, \sigma_{1}\right)+(1-\pi_{1}) \mathcal{N}\left( X | \mu_{2}, \sigma_{2}\right).$ Given data sampled from the mixture the goal is  to recover the true parameters $\left\{ \pi^\star_{k}, \mu^\star_{k}, \sigma^\star_{k}\right\}^{2}_{k=1}$. In particular we focus on learning the means and observe the singularity at $\mu_{1} = \mu_{2}$. In this case  $\{\pi_1, \pi_2,\sigma_{1 } = \sigma_{2}\}$ are unidentifiable. We now define the \emph{relative reparameterization} as $\vartheta =\{\mu_1,\mu_2\}\rep \{\mu,\Delta,\} = \Lambda$ characterize the convergence, especially taking the singularity into consideration, which is consistent with the empirically observed attractor behavior \cite{SingularDynamics, SingularDynamics2,SingularDynamica3} and illustrated in Fig.~\ref{fig: convergence}.

\begin{theorem}[\textbf{Convergence rate for GD under $\Delta\in\mathbb{R}$}]\label{lem: GD standard}\textbf{$\vartheta$-parameterization:} assume GD with fixed step size $\eta\in(0,2/L)$ and initialized sufficiently close to $\vartheta^\star$. If $\mu_1^\star\neq\mu_2^\star$, then $\|\vartheta^{(t)}-\vartheta^\star\|=O(\rho^t),\quad 0<\rho<1.$ If $\mu_1^\star=\mu_2^\star$, then no uniform linear rate holds and the convergence degenerates as $\|\vartheta^{(t)}-\vartheta^\star\|=O((1-c(\mu_{1}^{(t)}-\mu_{2}^{(t)})^2)^t)$ with constant $c>0$.

\noindent\textbf{$\Lambda$-parameterization:} let $\Delta\in\mathbb{R}$. If initialized sufficiently close to $\Lambda^\star$, then for both $\Delta^\star\neq0$ and $\Delta^\star=0$, $\|\Lambda^{(t)}-\Lambda^\star\|=O(\rho^t), 0<\rho<1,$ with $\rho$ uniformly bounded away from one.
\end{theorem}

We can now show that enforcing \emph{ordering} in the reparameterized model $(\Delta\geq 0)$ changes the convergence speed again. We consider two algorithmic approaches to enforce the constraint: firstly projected GD step $\Pi_{[0,\infty)}(\cdot)$ and secondly optimizing over $\Delta^2$ to ensure $\Delta\geq 0$.

\begin{theorem}[\textbf{Convergence rate for GD under $\Delta\in\mathbb{R}_+$}]\label{lem: GD projecte} Let us consider the two following cases:

\textbf{Projected GD:} Under $\Lambda$-parameterization and apply projected GD s.t. $\mu^{(t+1)}=\mu^{(t)}-\eta\,\partial_\mu \mathcal{L}(\Lambda^{(t)})$ and $\Delta^{(t+1)}=\Pi_{[0,\infty)}\left(\Delta^{(t)}-\eta\,\partial_\Delta \mathcal{L}(\Lambda^{(t)})\right).$ If $\Delta^\star>0$, then $\|\Lambda^{(t)}-\Lambda^\star\|=O(\rho^t),\quad 0<\rho<1.$ If $\Delta^\star=0$, then $\mu^{(t)}$ converges linearly while $\Delta^{(t)}=O(t^{-1}).$

\noindent\textbf{Optimize over $\Delta^2$:} Under $\Lambda$-parameterization with $\delta\ge0$, and apply GD s.t. $\mu^{(t+1)}=\mu^{(t)}-\eta\,\partial_\mu \mathcal{L}(\mu^{(t)},\delta^{(t)})$ and $\delta^{(k+1)}=\delta^{(t)}-\eta\,\partial_\delta \mathcal{L}(\mu^{(t)},\delta^{(t)}).$ If $\delta^\star>0$, convergence is linear. If $\delta^\star=0$, then $\delta^{(t)}=O(t^{-2})$ and $\Delta^{(t)}=\sqrt{\delta^{(t)}}=O(t^{-1}).$\end{theorem}

We can now move to Natural Gradient Descent (NGD) and observe a change in the behavior of the $\Theta$-parameterization.

\begin{theorem}[\textbf{Convergence rate for NGD under $\Delta\in\mathbb{R}$}]\label{lem FIM standard}
\textbf{$\vartheta$-parameterization:}
Under NGD, if $\mu_1^\star\neq\mu_2^\star$, then $\|\vartheta^{(t)}-\vartheta^\star\|=O(\rho^t),\quad 0<\rho<1.$ As $\mu_1^\star\to\mu_2^\star$, the contraction factor approaches one and uniform linear convergence fails.

\noindent\textbf{$\Lambda$-parameterization:} under NGD let $\Delta\in\mathbb{R}$  satisfies $\|\Lambda^{(t)}-\Lambda^{\star}\|=O(\rho^t),\quad 0<\rho<1,$ both away from and at $\Delta^\star=0$, with $\rho$ uniformly bounded away from one.\end{theorem}

\begin{remark}[Behavior \emph{near} the singularity.]Near but not exactly at the singularity, the same Hessian-based analysis (Theorem~\ref{lem: GD standard}) applies but with constants that now depend explicitly on the distance to the singular set. In the $\vartheta$-parameterization of the 2-GMM, when $|\mu_1-\mu_2|=\varepsilon>0$ is small, the smallest nonzero eigenvalue of the Hessian and of the FIM scales as $\Theta(\varepsilon^2)$. Consequently, (N)GD methods still converge linearly, but with a contraction factor of the form $1-c\varepsilon^2$ for some $c>0$, so that the number of iterations required to reach a fixed accuracy grows as $\Theta(\varepsilon^{-2})$. In contrast, in the reparameterized models, the Hessian in the $(\mu,\Delta)$ coordinates remains uniformly well conditioned as $\varepsilon\to 0$, and the linear rate persists with constants independent of $\varepsilon$. 
This clarifies that the apparent discrepancy between convergence far from and near the singularity is only quantitative: away from the singularity all parameterizations exhibit linear convergence of the same order, while sufficiently close to the singularity the standard parameterization suffers an arbitrarily slow linear rate due to vanishing curvature, whereas the reparameterized models maintain uniform geometric contraction until the singular point itself is reached.\end{remark}

\begin{remark}[EM Algorithm]\label{rem: EM}For learning GMMs, a common approach is to consider an EM based approach \cite{EMOriginal,EMNew} where the main qualitative conclusion is that its local convergence behavior is fundamentally governed by the same identifiability structure as gradient-based methods, and is therefore largely invariant to smooth reparameterization. In both the standard parameterization $(\mu_1,\mu_2)$ and the reparameterized form $(\mu,\Delta)$ with $\mu_2=\mu+\Delta$, following \cite{Zhao2018StatisticalCO,Dwivedi2018SingularityMA} we see that EM converges locally at a linear rate whenever the mixture components are well separated. Far away from the singularity, the Jacobian of the EM update map at the fixed point has spectral radius strictly smaller than one, yielding geometric convergence with constants independent of the parameterization. Near but not at the singularity, when the separation is $\varepsilon>0$ small, EM remains linearly convergent but with a contraction factor of the form $1-\Theta(\varepsilon^{2})$, so that convergence becomes arbitrarily slow as $\varepsilon\to 0$. At the singularity itself, EM loses uniformly contract in the non-identifiable direction in both parameterizations.
\end{remark}

\section{Neural Networks}\label{sec: NN}

Singularities have been observed previously in the NN context \cite{SingularDynamics,eliminationSingularity,SkipConnection,InitalConePaper,SingularDynamica3,LinearSingularity}, in the following we explicitly characterize the dynamics under consideration of singularities and relative reparameterization.
We consider the two–unit neural network
$\mathbb{R}\rightarrow\mathbb{R}:f(x;u_1,u_2,w_1,w_2)=u_1\sigma(w_1 x)+u_2\sigma(w_2 x)$, where $\sigma(\cdot)$ is a twice continuously differentiable activation with nonvanishing derivative on a set of positive measure, and the population risk is the squared loss with respect to a realizable target $f^\star(x)=u^\star\sigma(w^\star x)$. This model is overparameterized and admits a singular manifold of global minimizers given by ${u_1+u_2=u^\star,\ w_1=w_2=w^\star}$. A reference-based reparameterization,  is obtained wlog. by taking $(u_1,w_1)$ as reference parameters and defining difference parameters $(\Delta_u,\Delta_w)$ via $u_2=u_1+\Delta_u$ and $w_2=w_1+\Delta_w$. The singularity corresponds to $(\Delta_u,\Delta_w)=(0,0)$, where the two hidden units coincide and the parameterization becomes non-identifiable. 
We can formally derive the convergence rate as follows and numerically illustrate it in Fig.~\ref{fig: convergence}.

\begin{theorem}[\textbf{Convergence rate for GD under $\Delta\in\mathbb{R}$}]
\label{lem: NN standard}
\noindent\textbf{~$\vartheta$-parameterization:} under GD with a sufficiently small constant $\eta$, if $w_1^\star\neq w_2^\star$, then the iterates converge locally linearly to a minimizer, with rate $O(\rho^t),\ \rho\in(0,1)$. If $w_1^\star=w_2^\star$, then linear convergence is not uniform: near the singularity, when $\varepsilon:=|w_1-w_2|$ is small but nonzero, the contraction factor is of the form $1-c\varepsilon^2$. The convergence along the difference direction becomes arbitrarily slow as $\varepsilon\to 0$.

\noindent\textbf{$\Lambda$-parameterization:} under GD for sufficiently small $\eta$, the iterates converge locally linearly to a minimizer both away from and at the singularity $(\Delta_u^\star,\Delta_w^\star)=(0,0)$, with rate $O(\rho^t)$ for some $\rho\in(0,1)$ independent of the distance to the singularity.
\end{theorem}

\section{Properties of the Fisher Information Matrix}\label{sec: FIM}

In general, the FIM provides a Riemannian metric however the FIM degenerates at the singularity as the parameter space is no longer a Riemannian manifold at this point which has been studied previously (e.g. \cite{NGD1,watanabe_2009}).
There are several applications in statistics, one of the most important being the Cramer-Rao bound \cite{Cramer-Rao,Cramer-Rao-Frank}. We can now use it to characterize the geometry of the model under reparameterization.

\noindent\textbf{Covariant Aspects.} In information geometry, the reparameterization is seen as a change of coordinates on a Riemannian manifold (assuming the model is not singular), and the intrinsic properties of curvature are unchanged under different parameterization.

\begin{theorem}[\textbf{Covariance of the FIM under Reparameterization}]\label{lem: covariant} 
Let ${\psi}$ and ${\phi}$ be $k$-vectors which parameterize an estimation problem, and suppose that ${\psi}$ is a continuously differentiable function of ${\phi}$, then, $\mathcal{I}_{\phi}({\phi})={J}^{\top}\mathcal{I}_{\psi}({\psi}({\phi}))\,{J}$ where the $(i,j)$-th element of the $k\times k$ Jacobian matrix $J$ is defined by $J_{ij}=\frac{\partial \psi_i}{\partial \phi_j}$ \cite{TheoryPointEstimation}.
\end{theorem}

This covariance explains a key observation from the earlier convergence analysis. For the GMM in the $\vartheta$-parameterization, the FIM degenerates as $\mu_1\to\mu_2$, because the Jacobian mapping from the identifiable quotient coordinates to $(\mu_1,\mu_2)$ becomes rank-deficient. Consequently, both the FIM and the Hessian of the log-likelihood develop small eigenvalues, which directly slow GD through poor conditioning. In contrast, the $\Lambda$-reparameterization aligns coordinates with identifiable and non-identifiable directions. In this coordinate system, the Jacobian absorbs the degeneracy, and the FIM in the $\Delta$-direction remains well-scaled until $\Delta=0$. Thus, the apparent improvement in convergence rate under reparameterization does not contradict covariance; it reflects a change in coordinate conditioning rather than a change in intrinsic geometry.

\noindent\textbf{Invariant Aspects.} This invariance explains why NGD exhibited parameterization-independent convergence behavior in the earlier analysis. NGD follows the steepest descent direction with respect to the Fisher metric, which depends only on the intrinsic Riemannian geometry, not on the chosen coordinates. Therefore, once distances are measured in Fisher length, optimization trajectories are invariant under reparameterization. 

\begin{theorem}[\textbf{Invariance of the Length Element of the FIM under Reparameterization}]\label{lem: length element} The length elements $\ds^2$ of the FIM coincide under reparameterization. Therefore for $\psi$- or $\phi$-coordinate system the following holds: $\ds^2(\psi)=\Delta\psi \nabla^2 F(\psi) \Delta\psi~=~\Delta\phi \nabla^2 F^2(\phi) \Delta\phi=\ds^2(\phi).$ \end{theorem}

In the GMM setting, this implies that the slow convergence observed near singularities under standard GD is not an intrinsic property of the statistical model but rather a consequence of measuring distances and gradients in a Euclidean metric that is misaligned with the manifold geometry. NGD corrects for this by rescaling updates according to the FIM, effectively neutralizing the coordinate-induced ill-conditioning caused by the singularity.

\section{Discussion and Related Work}
Learning dynamics have been studied for a wide range of machine learning models.
While the proofs in this paper build on existing approaches  \cite{poliak1987introduction,Nesterov04,Lee2016GradientDO,Zhao2018StatisticalCO,Dwivedi2018SingularityMA,MatrixManifolds}, they have now been considered explicitly to study the relative reparameterization of singular models.
Parameter ordering has been practically used in Markov Chain Monte Carlo for GMM \cite{FiniteMM} and is applied for EM on GMM, where ordering is usually (implicitly) assumed but not enforced in the optimization \cite{FiniteMM,FiniteMixtureModels}. We extend this line of research by explicitly analyzing the influence of parameter ordering through reparameterization on the learning dynamics. 
In Section~\ref{sec: NN} we show that this approach is also relevant for modern approaches such as NN. While works on reparameterization of neural networks exist \cite{sharpminima}, they do not approach it from the perspective of learning dynamics in the context of model identifiability questions. In the context of information criteria \cite{Watanabe2010, watanabe_2009,Watanabe2013} introduces extensions to singular models by blowing the space up into higher dimensions. Keeping the parameter dimension the same, \cite{esser_OPT_towards} proposes to consider smooth manifold approximations of the singular space to handle singularities. Finally, using that  in singular spaces a majority of the space is still smooth \cite{Sun_2019_Lightlike,Lin_2019_Tractable} proposes local analysis of the FIM.

\emph{In conclusion, in this paper we provide a concise overview of the behavior of learning dynamics under relative reparameterization. The apparent improvement in convergence rates under reparameterization in GMMs and NNs is fully consistent with information-geometric principles. Reparameterization improves first-order optimization not by altering the underlying geometry, but by aligning Euclidean updates with the intrinsic Riemannian structure of the statistical manifold, particularly near singularities where identifiability breaks down.}

\printbibliography

\appendix
\section{Proofs}
In all cases, the convergence proofs follow the same general, established structure (e.g. \cite{poliak1987introduction,Nesterov04,Lee2016GradientDO,MatrixManifolds}): one studies the local error recursion obtained by linearizing the update map around a stationary point and expresses it in terms of the Hessian (or its preconditioned version). Writing the iterate as $\psi^{(t+1)}-\psi^\star = (I-\eta \tilde H)(\psi^{(t)}-\psi^\star)+o(|\psi^{(t)}-\psi^\star|)$, where $\tilde H$ is either the Hessian $\nabla^2 \mathcal{L}(\psi^\star)$ for GD or a similarity transform such as $\mathcal I(\psi^\star)^{-1}\nabla^2 \mathcal{L}(\psi^\star)$ for NGD methods, reduces convergence to a spectral analysis of $\tilde H$.  If the Hessian is PD with eigenvalues bounded away from zero and infinity, the iteration matrix has spectral radius strictly smaller than one and yields linear convergence. Slow or sublinear convergence arises precisely when the Hessian (or FIM) becomes ill-conditioned or degenerate, as in the presence of singularities or non-identifiability: small eigenvalues flatten the local quadratic approximation, weaken the contraction, and dominate the asymptotic rate. Reparameterization changes the coordinate representation of the Hessian but not its intrinsic geometry; however, it can remove degeneracies or align coordinates with identifiable directions, thereby restoring uniform spectral bounds and faster convergence.

\emph{Proof Th.~\ref{lem: GD standard}}
When $\mu_1^\star\neq\mu_2^\star$, the Hessian $\nabla^2\mathcal{L}(\vartheta^\star)$ is PD with eigenvalues bounded away from zero, so standard smooth strongly convex analysis yields linear convergence. At $\mu_1^\star=\mu_2^\star$, the model is non-identifiable and the smallest eigenvalue of the Hessian scales as $(\mu_1-\mu_2)^2$. The linearized error recursion thus has a contraction factor approaching one as $\mu_{1}^{(t)}-\mu_{2}^{(t)}\to0$, implying arbitrarily slow convergence and ruling out a uniform linear rate.  In the $\Lambda$ coordinates, the separation parameter $\Delta$ directly captures identifiability. A local Taylor expansion shows that $\partial^2_{\Delta\Delta}\mathcal{L}(\Lambda^\star)>0$ even at $\Delta^\star=0$, and the full Hessian is PD with bounded condition number. Hence $L$ is locally smooth and strongly convex, and the GD recursion yields linear convergence uniformly across both regimes. \qed

\emph{Proof Th.~\ref{lem: GD projecte}}
When $\Delta^\star>0$, the constraint is inactive in a neighborhood of the optimum and the result follows from Th.~2. When $\Delta^\star=0$, the constraint is active and the projection introduces a nonsmooth boundary. Expanding $\partial_\Delta \mathcal{L}(\Lambda)=c\Delta+O(\Delta^2)$ near $\Delta=0$ yields the effective recursion $\Delta^{(t+1)}\approx\max\{0,\Delta^{(t)}-\eta c(\Delta^{(t)})^2\}$, whose solution decays as $O(t^{-1})$. The $\mu$-coordinate remains unconstrained and strongly convex, hence converges linearly. 
For $\delta^\star>0$, the map $\delta\mapsto\sqrt{\delta}$ is smooth and the Hessian is well conditioned, implying linear convergence. At $\delta^\star=0$, the chain rule gives $\partial_\delta L=(2\sqrt{\delta})^{-1}\partial_\Delta L$, so the effective curvature diverges as $\delta\to0$. The resulting recursion behaves as $\delta^{(t+1)}=\delta^{(t)}-\eta c+O(\eta\sqrt{\delta^{(t)}})$, which solves to $\delta^{(t)}=O(t^{-2})$, yielding sublinear decay in $\Delta$. \qed

\emph{Proof Th.~\ref{lem FIM standard}}
Away from the singularity, the FIM is PD and uniformly well conditioned, so the linearized natural gradient operator has eigenvalues strictly inside the unit circle. At the singularity, the FIM degenerates due to non-identifiability, and its inverse amplifies curvature errors, causing the spectral radius of the update to approach one.
In these $\Lambda$-coordinates, the FIM remains nondegenerate in the identifiable directions even at $\Delta=0$. The linearized update is a similarity transform of the intrinsic FIM/Hessian operator on the quotient manifold, whose eigenvalues are bounded away from zero and infinity. Consequently, the spectral radius of the update matrix is uniformly $<1$, yielding linear convergence in all cases. \qed

\emph{Proof Th.~\ref{lem: NN standard}} Follows directly as the proof of Th.~\ref{lem: GD standard}. \qed

\emph{Proof Th.~\ref{lem: covariant}} Assume a function $f$, parameterized by $\psi$, and a  function $g$, with parameters $\phi$. Furthermore, there is a reparameterization function $\xi(\psi) = \phi$. From there, the reparameterization of the FIM is given by: $\gI_{i j}^{\psi}(\psi)= \mathbb{E}\big[\big(\frac{\partial\log f(x | \psi)}{\partial \psi_{i}} \big)\big(\frac{\partial\log f(x | \psi)}{\partial \psi_{j}} \big)\big] =\mathbb{E}\big[\big(\frac{\partial\log g(x | \xi(\psi))}{\partial \theta_{i}} \big)\big(\frac{\partial \log g(x | \xi(\psi))}{\partial \theta_{j}}\big)\big]  =\mathbb{E}\big[\big(\sum_{k} \frac{\partial \log (x | \phi)}{\partial \eta_{k}} \frac{\partial \eta_{k}}{\partial \theta_{i}}\big)\cdot \big(\sum_{l} \frac{\partial \log g(x | \phi)}{\partial \eta_{l}} \frac{\partial \eta_{l}}{\partial \theta_{j}}\big)\big] $therefore $ \gI^{\psi}(\psi) =(J_{\psi}^{\phi})^\top \gI^{\phi}(\phi) J_{\psi}^{\phi} $ \qed

\emph{Proof Th.~\ref{lem: length element}} We see this directly by considering the following. Let $F(\psi)$ be the log-normalizer of an exponential family, and $F^*(\phi)$ the negentropy convex conjugate. The FIM can be expressed either in the $\psi$- or $\phi$-coordinate system: $\I_\psi(\psi)=\nabla^2F(\psi),\quad \I_\phi(\phi)=\nabla^2F^*(\phi).$ The FIM is not invariant, it is covariant: That is, $\I_\phi(\psi(\phi))\not= \I_\phi(\phi)$ since $\psi(\phi)=\nabla F^*(\phi)$. However, the Riemannian length elements $\ds^2(\psi)=\Delta\psi \nabla^2 F(\psi) \Delta\psi$ and $\ds^2(\phi)=\Delta\phi \nabla^2 F^2(\phi) \Delta\phi$ coincide: The length element is invariant by reparameterization. Indeed, we have $\Delta\psi= \nabla^2 F^*(\phi)\Delta\phi$ ($\psi$ is a contravariant vector and $\phi$ is a covariant vector, and we use the metric tensor to raise/lower indices). Thus, we have $\ds^2(\psi)=\Delta\phi^\top \nabla^2 F^*(\phi)\nabla^2 F(\psi)\nabla^2 F^*(\phi)\Delta\phi$ Since $\nabla^2 F^*(\phi)\nabla^2 F(\psi)=\I$ (Crouzeix identity), we have $\ds^2(\psi)=\Delta\phi\nabla^2 F^*(\phi)\Delta\phi=\ds^2(\phi)$. \qed

\end{document}